\pdfoutput=1

\documentclass[11pt]{article}

\usepackage[]{acl}

\usepackage{times}
\usepackage{latexsym}

\usepackage[T1]{fontenc}

\usepackage[utf8]{inputenc}

\usepackage{microtype}

\usepackage{graphicx}
\usepackage{amsmath}

\title{The Document Vectors Using Cosine Similarity Revisited}

\usepackage{amssymb}  
\usepackage{ifsym}

\author{Zhang Bingyu$^{\triangle}$ \\
  \And Nikolay Arefyev$^{\diamondsuit,\nabla,\triangle}$ \\
  \AND
  \vspace{-0.8cm}\\
  $^{\triangle}$National Research University Higher School of Economics / Moscow, Russia \\
  $^\diamondsuit$Samsung Research Center Russia / Moscow, Russia \\
  $^{\nabla}$Lomonosov Moscow State University / Moscow, Russia \\
  \texttt{bchzhan\_1@edu.hse.ru,} \texttt{nick.arefyev@gmail.com} \\
}

\begin{document}
\maketitle
\begin{abstract}
The current state-of-the-art test accuracy (97.42\%) on the IMDB movie reviews dataset was reported by \citet{thongtan-phienthrakul-2019-sentiment} and achieved by the logistic regression classifier trained on the Document Vectors using Cosine Similarity (DV-ngrams-cosine) proposed in their paper and the Bag-of-N-grams (BON) vectors scaled by Naive Bayesian weights. While large pre-trained Transformer-based models have shown SOTA results across many datasets and tasks, the aforementioned model has not been surpassed by them, despite being much simpler and pre-trained on the IMDB dataset only. 

In this paper, we describe an error in the evaluation procedure of this model, which was found when we were trying to analyze its excellent performance on the IMDB dataset. We further show that the previously reported test accuracy of 97.42\% is invalid and should be corrected to 93.68\%. We also analyze the model performance with different amounts of training data (subsets of the IMDB dataset) and compare it to the Transformer-based RoBERTa model. The results show that while RoBERTa has a clear advantage for larger training sets,
the DV-ngrams-cosine performs better than RoBERTa when the labelled training set is very small (10 or 20 documents). Finally, we introduce a sub-sampling scheme based on Naive Bayesian weights for the training process of the DV-ngrams-cosine, which leads to faster training and better quality. 
\end{abstract}

\section{Introduction}

The word2vec algorithm originally published by \citet{mik_wordvec2013} is among the most famous methods to train vector representations of words. Soon after the emergence of word2vec, a similar method to build vector representations of documents was originally proposed by \citet{mik_docvec2014} and further studied by \citet{mes_2015}. It is known under different names, including Paragraph Vectors, Sentence Vectors, doc2vec, etc. This method jointly learns word embeddings and document embeddings such that a binary classifier can predict if a given word occurs in a particular document given only the corresponding embeddings. More formally, the following objective is minimized:
\begin{equation}
    \sum_{d\in D}\sum_{w\in W_d} [-\log\sigma (v_d^T v_w) - \sum_{w' \sim V} \log\sigma(-v_d^T v_{w'})]
\end{equation}
Here $D$ denotes the set of documents, $W_d$ is the list of words that make up the document $d$, $w'$ is a word randomly sampled from the full vocabulary $V$, also known as a negative sample~\citep{Goldberg_2014}. Finally, $v_d$ and $v_w$ are the learnt embeddings of $d$ and $w$. Intuitively, for each document, an embedding is learnt that has high similarity to the embeddings of those words that occur in this document and low similarity to the embeddings of some random words.

Later \citet{li_2016} switched from single words to n-grams and observed significant improvements. Building on that, \citet{thongtan-phienthrakul-2019-sentiment} studied different objective functions. They have found that the cosine similarity outperforms the dot product, which led to a modified model called the Document Vectors using Cosine Similarity (we will call it \textbf{DV-ngrams-cosine} for short). The new objective is:
\begin{equation}
\begin{aligned}
    \sum_{d\in D}&\sum_{u\in U_d} [-\log\sigma (\alpha cos(v_d, v_u))\\
    - &\sum_{u' \sim V} \log\sigma(-\alpha cos(v_d, v_{u'}))],
\end{aligned}
\end{equation}
where $U_d$ denotes the set of all n-grams in $d$, $v_u$ is the embedding of the n-gram $u$ from $d$, $v_{u'}$ is the embedding of a randomly sampled n-gram, and $\alpha$ is a hyperparameter.

In the same paper, the authors proposed an ensemble consisting of the document embeddings from DV-ngrams-cosine and the Bag-of-N-grams vectors scaled by Naive Bayesian weights (\textbf{NB-weighted BON} for short). They concatenated these two representations and trained the logistic regression classifier on top. The ensemble was reported to have very high test accuracy (97.42\%) on the IMDB movie reviews dataset (\citet{maas-etal-2011-learning}). To the best of our knowledge, this accuracy remains the SOTA result on IMDB. Even large Transformer-based models pre-trained on a huge amount of texts, both in-domain and out-of-domain, have shown lower accuracy on this dataset~\cite{yang2019xlnet,suchin2020,arefyev2021nb-mlm}.

This extraordinary performance of such a simple model motivated us to thoroughly study the model and its implementation trying to understand the reasons behind its success. Unfortunately, during this study, we found a bug in the implementation of the evaluation procedure of the ensemble, which had made the estimation of the accuracy incorrect.

In our paper, we re-evaluate the ensemble as well as its individual components. We
show that the originally reported test accuracy of the ensemble (97.42\%) is incorrect and shall be corrected to 93.68\%, which is only 0.55\% higher than the accuracy on pure DV-ngrams-cosine embeddings. 

Additionally, we analyze how the amount of training data affects the performance of the ensemble, as well as its individual components, and also the Transformer-based RoBERTa model~\cite{liu2020roberta}, which has recently shown SOTA or near-SOTA results over a variety of tasks and datasets. Surprisingly, we have observed that DV-ngrams-cosine outperforms RoBERTa when the number of labelled training examples is small (10 or 20). We also ensemble RoBERTa with DV-ngrams-cosine, but only have achieved a marginal improvement. Finally, we propose a modification for the training process of DV-ngrams-cosine that results in faster training and better accuracy. The code reproducing our experiments is publicly available~\footnote{https://github.com/Bgzh/dv\_cosine\_revisited}.

\section{Re-evaluation of the ensemble}
\label{sec:ens_n_problem}
In the aforementioned ensemble proposed by \citet{thongtan-phienthrakul-2019-sentiment}, the NB-weighted BON and the DV-ngrams-cosine are concatenated and fed into the logistic regression classifier. 
However, we have found that in the original implementation the two vectors concatenated to obtain a single training or test example usually correspond to two different documents of the same class (see details in Appendix~\ref{sec:appendix_bug}). Specifically, the DV-ngrams-cosine vectors and the BON vectors are built from two different files having different orders of examples. As a result, after the concatenation, each input to the logistic regression corresponds to a combination of two examples. Due to the special structure of the files, those examples are guaranteed to belong to the same class and the same subset. For instance, a positive example from the test set is concatenated with another positive example from the test set. 

In~Appendix~\ref{sec:subsec_hard_n_easy} we provide an analysis that shows the reasons of high performance of this concatenation of two representations. From this analysis it follows that most examples from IMDB are correctly classified with high confidence (a large logit) using any of two representations, i.e. they are easy examples. Less than 10\% of examples are classified incorrectly by each representation (hard examples), but they often obtain low confidence (a logit near zero). Hard examples are more often combined with  easy examples just because of their dominance. In these cases, the logit from the easy example often outweigh the logit from the hard one resulting in the correct final prediction.

Thus, in both the training and the test sets, hard examples are often combined with simpler examples, making the classification task easier. In this process, the knowledge of the true labels is implicitly exploited to combine the examples this way, in both training and testing. This leads to an incorrect estimation of the classification accuracy for future examples.

After fixing this issue, we have observed that the combination of different representations of the same document leads to the test accuracy of 93.68\% instead of 97.42\% originally reported. Compared to the pure DV-ngrams-cosine embeddings, the ensemble improves the test accuracy by 0.55\%, not 4.29\% reported previously. This improvement also better agrees with the improvements of less than 1\% observed by~\citet{li_2016} for similar ensembles with the predecessor model DV-ngram. As a sanity check, Appendix~\ref{sec:appendix_reev} additionally reports the accuracy for different schemes of combining the two representations, showing that higher accuracy can be achieved only by those schemes that exploit the knowledge of the test labels.\newline

\section{Further analysis of performance}
\label{sec:train_size}
In his section we further analyze the performance of the ensemble described above, comparing it to its individual components as well as to the recently introduced Transformer-based RoBERTa model~\cite{liu2020roberta}. We study the performance of these models depending on the number of labelled examples in the training set.
\begin{figure}[ht] 
    \centering 
    \includegraphics[width=0.45\textwidth]{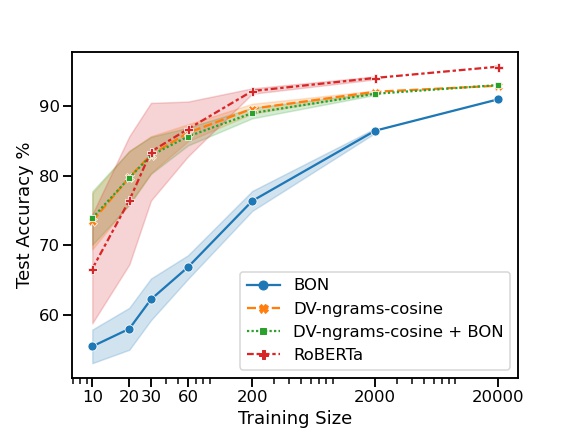} 
    \caption{The performance of different models on training sets of different sizes. The mean values and standard deviations were calculated over 10 random subsets for RoBERTa and 30 random subsets for other models for each training set size. BON in the legend implies NB-weighted BON.}
    \label{fig:train_size}
\end{figure}

For a more fair comparison, the most important hyperparameters of each model were tuned on the validation set, employing the train/validation/test split of the IMDB dataset provided by \citep{suchin2020}. Subsets of different sizes from 10 to 20000 examples were randomly sampled from the training set. The logistic regression classifier was trained on these subsets using the DV-ngram-cosine embeddings, the NB-weighted BON vectors, or their concatenation as its input representation. 

We tuned the L2-regularization strength $C$ of the classifier individually for each subset of the training set. Additionally, we multiplied the DV-ngram-cosine embeddings before concatenating them to the BON vectors in order to balance the magnitudes of the two representations, which may help the classifier to benefit from both representations. The scaling factor was also selected on the validation set. 

The pre-trained RoBERTa base model\footnote{\url{https://pytorch.org/hub/huggingface_pytorch-transformers/}} was fine-tuned on a part (10 out of 30) of the same subsets of the training set, using the validation set for early stopping. We used a batch size of 32, with a maximum learning rate of 1e-5, recommended by fairseq\footnote{\url{https://github.com/pytorch/fairseq/blob/main/examples/roberta/README.custom_classification.md}}.

As shown in Fig.~\ref{fig:train_size}, the fine-tuned RoBERTa model usually achieves higher test accuracy. But when the number of labelled training examples is very small (10 or 20), the logistic regression on the DV-ngrams-cosine embeddings shows higher mean test accuracy and lower standard deviation. This result corroborated the notion that small models can be a better choice when the data are scarce.

On the other hand, logistic regression on the BON vectors performs significantly worse than all other models across all training set sizes. Finally, we don't observe any significant improvements from the ensembling when the training set size is less than 20k, as the difference is within one standard deviation.

It is important to notice that the DV-ngrams-cosine embeddings were pre-trained on the in-domain examples from the whole IMDB dataset, while RoBERTa was pre-trained on a huge but general-domain corpus.
It is likely that the domain adaptation techniques~\cite{suchin2020} will help RoBERTa when the number of labelled examples is small. However, for our study, we decided to compare the most standard approaches to training the corresponding models.

\section{NB Sub-Sampling}
\label{sec:subsamp}
In this section, we improve the training procedure of DV-ngrams-cosine by applying a sub-sampling procedure based on the Naive Bayesian weights of ngrams (\textbf{NB Sub-Sampling}) in order to make the model focus more on sentiment-related ngrams while building the document embeddings.

\begin{figure}[!htb] 
    \centering 
    \includegraphics[width=0.47\textwidth]{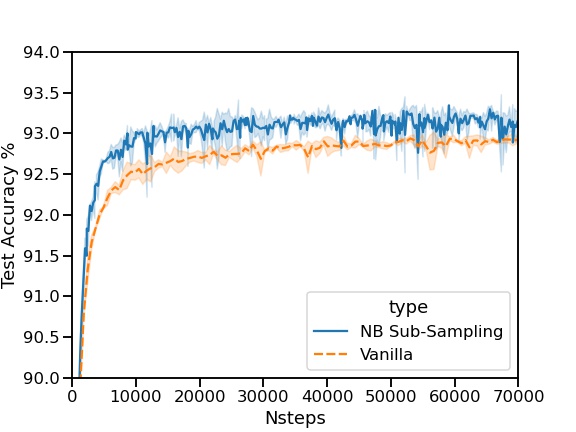}
    \caption{Training process with and without NB sub-sampling. The test accuracy of the logistic regression built on top of the document vectors is plotted. The mean values and standard deviations were calculated over 3 runs for each type.}
    \label{fig:sub_samp}
\end{figure}

\begin{table*}
\centering
\begin{tabular}{lc}
\hline
\textbf{Model} & \textbf{Test Accuracy \%}\\
\hline
\textit{Models trained on the original training set of IMDB (25K)} &  \\
\textbf{NB-weighted BON} & 91.29 \\
\textbf{DV-ngrams-cosine} & 93.13 \\
\textbf{DV-ngrams-cosine + NB-weighted BON \cite{thongtan-phienthrakul-2019-sentiment}} & $\#$97.42 \\
\textbf{DV-ngrams-cosine + NB-weighted BON (re-evaluated)} & 93.68 \\
\hline
\textit{Models trained using the train/dev split from~\cite{suchin2020} (20K/5K)} & \\
\textbf{DV-ngrams-cosine with NB sub-sampling} & 93.36 \\
\textbf{RoBERTa} & 95.79 \\
\textbf{DV-ngrams-cosine + RoBERTa} & 95.92 \\
\textbf{DV-ngrams-cosine with NB sub-sampling + RoBERTa} & 95.94 \\
\hline
\end{tabular}
\caption{Test results on the IMDB dataset. $\#$ indicates incorrect previously reported results.}
\label{tab:imdb_acc}
\end{table*}

Inspired by the previous works (\citet{wang-manning-2012-baselines}, \citet{arefyev2021nb-mlm}), we trained a multinomial Naive Bayesian Classifier and exploited its weights to calculate the importance of each ngram $f_i$ for the final classification task:
\begin{equation}
\label{eq:nbw}
h_i = |\log p(f_i|y=1) - \log p(f_i|y=0)|
\end{equation}
In each epoch we put an ngram into training with the probability
\begin{equation}
\label{eq:ss}
    p(f_i) = min(\exp (h_i / n_a) / n_b, 1),
\end{equation}
where $n_a$ and $n_b$ are the hyperparameters. The choices are purely empirical. We tried different combinations of $n_a$ and $n_b$ and found 2 and 3 (respectively) to be the best in them.

The comparison of the training process with and without NB sub-sampling is shown in Fig. \ref{fig:sub_samp} (refer to Appendix~\ref{sec_appendix_subsamp} for details of the experiments and the accuracy on the validation set).

The runs with NB sub-sampling progress faster and show a distinct advantage after 2500 steps. After 30k steps, the runs with NB sub-sampling stagnated and kept fluctuating in a small region; the vanilla runs stagnated after 50k steps, in a lower area. It is also worth noticing that although the labels of the training set are used during pre-training for sub-sampling, we did not observe any significant overfitting due to that. Neither the validation score nor the test score showed a tendency to decay long after reaching the plateau, indicating that this sub-sampling scheme can be used as an add-on to the original model, boosting its performance while not creating additional overfitting trouble.

\section{Ensemble DV-ngrams-cosine and RoBERTa}
The ensemble proposed in (\citet{thongtan-phienthrakul-2019-sentiment}) and described in Section~\ref{sec:ens_n_problem} combines two different representations of documents, which are the DV-ngrams-cosine embeddings and the NB-weighted BON vectors. However, we have observed in Section~\ref{sec:train_size} that the BON vectors are quite weak on their own, while RoBERTa outperforms all other models unless the number of examples is very small. Thus, it is interesting if DV-ngram-cosine can help RoBERTa. 
In this section, we combine the DV-ngrams-cosine (with or without NB sub-sampling) with the output of the last hidden layer of RoBERTa, and test on the IMDB dataset. Again, the train/validation/test splits by \citet{suchin2020} were used. A scaling factor on the DV-ngrams-cosine and the hyperparameter $C$ in the logistic regression were tuned on the validation set.\newline
The results are shown in Table~\ref{tab:imdb_acc}. Although RoBERTa is a much stronger model than DV-ngram-cosine, combining them has shown a small improvement of 0.13-0.15\%.

\section{Conclusion}
The ensemble featuring the DV-ngrams-cosine reported by \citet{thongtan-phienthrakul-2019-sentiment} was re-evaluated. The test accuracy of this ensemble on the IMDB dataset was corrected from 97.42\% to 93.68\%. The DV-ngrams-cosine embeddings with the logistic regression on top were compared with RoBERTa using different amounts of training data. In this comparison, the DV-ngrams-cosine has surprisingly outperformed RoBERTa for a small number of training examples (10 or 20 documents). A sub-sampling scheme based on the Naive Bayesian weights was introduced to the training process of the DV-ngrams-cosine, resulting in faster training and better quality.

\section*{Acknowledgements}
We are grateful to our anonymous reviewers. This research was partially supported by the Basic Research Program at the HSE University.

\bibliography{anthology,custom}

\begin{thebibliography}{12}
\expandafter\ifx\csname natexlab\endcsname\relax\def\natexlab#1{#1}\fi

\bibitem[{Arefyev et~al.(2021)Arefyev, Kharchev, and
  Shelmanov}]{arefyev2021nb-mlm}
Nikolay Arefyev, Dmitry Kharchev, and Artem Shelmanov. 2021.
\newblock Nb-mlm - efficient domain adaptation of masked language models for
  sentiment analysis.
\newblock \emph{EMNLP}, pages 9114--9124.

\bibitem[{Goldberg and Levy(2014)}]{Goldberg_2014}
Yoav Goldberg and Omer Levy. 2014.
\newblock word2vec explained: deriving mikolov et al.'s negative-sampling
  word-embedding method.
\newblock \emph{CoRR}.

\bibitem[{Le and Mikolov(2014)}]{mik_docvec2014}
V.~Quoc Le and Tomas Mikolov. 2014.
\newblock Distributed representations of sentences and documents.
\newblock \emph{ICML}, pages 1188--1196.

\bibitem[{Li et~al.(2015)Li, Liu, Du, Zhang, and Zhao}]{li_2016}
Bofang Li, Tao Liu, Xiaoyong Du, Deyuan Zhang, and Zhe Zhao. 2015.
\newblock Learning document embeddings by predicting n-grams for sentiment
  classification of long movie reviews.
\newblock \emph{CoRR}.

\bibitem[{Liu et~al.(2020)Liu, Ott, Goyal, Du, Joshi, Chen, Levy, Lewis,
  Zettlemoyer, and Stoyanov}]{liu2020roberta}
Yinhan Liu, Myle Ott, Naman Goyal, Jingfei Du, Mandar Joshi, Danqi Chen, Omer
  Levy, Mike Lewis, Luke Zettlemoyer, and Veselin Stoyanov. 2020.
\newblock \href {https://openreview.net/forum?id=SyxS0T4tvS} {Ro{\{}bert{\}}a:
  A robustly optimized {\{}bert{\}} pretraining approach}.

\bibitem[{Maas et~al.(2011)Maas, Daly, Pham, Huang, Ng, and
  Potts}]{maas-etal-2011-learning}
Andrew~L. Maas, Raymond~E. Daly, Peter~T. Pham, Dan Huang, Andrew~Y. Ng, and
  Christopher Potts. 2011.
\newblock \href {https://aclanthology.org/P11-1015} {Learning word vectors for
  sentiment analysis}.
\newblock In \emph{Proceedings of the 49th Annual Meeting of the Association
  for Computational Linguistics: Human Language Technologies}, pages 142--150,
  Portland, Oregon, USA. Association for Computational Linguistics.

\bibitem[{Mesnil et~al.(2015)Mesnil, Mikolov, Ranzato, and Bengio}]{mes_2015}
Grégoire Mesnil, Tomas Mikolov, Marc'Aurelio Ranzato, and Yoshua Bengio. 2015.
\newblock Ensemble of generative and discriminative techniques for sentiment
  analysis of movie reviews.
\newblock \emph{international conference on learning representations}.

\bibitem[{Mikolov et~al.(2013)Mikolov, Chen, Corrado, and
  Dean}]{mik_wordvec2013}
Tomas Mikolov, Kai Chen, Greg Corrado, and Jeffrey Dean. 2013.
\newblock Efficient estimation of word representations in vector space.
\newblock \emph{CoRR}.

\bibitem[{Suchin et~al.(2020)Suchin, Ana, Swabha, Kyle, Iz, Doug, and
  A.}]{suchin2020}
Gururangan Suchin, Marasović Ana, Swayamdipta Swabha, Lo~Kyle, Beltagy Iz,
  Downey Doug, and Noah~Smith A. 2020.
\newblock Don't stop pretraining: Adapt language models to domains and tasks.
\newblock \emph{ACL}, pages 8342--8360.

\bibitem[{Thongtan and
  Phienthrakul(2019)}]{thongtan-phienthrakul-2019-sentiment}
Tan Thongtan and Tanasanee Phienthrakul. 2019.
\newblock \href {https://doi.org/10.18653/v1/P19-2057} {Sentiment
  classification using document embeddings trained with cosine similarity}.
\newblock In \emph{Proceedings of the 57th Annual Meeting of the Association
  for Computational Linguistics: Student Research Workshop}, pages 407--414,
  Florence, Italy. Association for Computational Linguistics.

\bibitem[{Wang and Manning(2012)}]{wang-manning-2012-baselines}
Sida Wang and Christopher Manning. 2012.
\newblock \href {https://aclanthology.org/P12-2018} {Baselines and bigrams:
  Simple, good sentiment and topic classification}.
\newblock In \emph{Proceedings of the 50th Annual Meeting of the Association
  for Computational Linguistics (Volume 2: Short Papers)}, pages 90--94, Jeju
  Island, Korea. Association for Computational Linguistics.

\bibitem[{Yang et~al.(2019)Yang, Dai, Yang, Carbonell, Salakhutdinov, and
  Le}]{yang2019xlnet}
Zhilin Yang, Zihang Dai, Yiming Yang, G.~Jaime Carbonell, Ruslan Salakhutdinov,
  and V.~Quoc Le. 2019.
\newblock Xlnet: Generalized autoregressive pretraining for language
  understanding.
\newblock \emph{ADVANCES IN NEURAL INFORMATION PROCESSING SYSTEMS 32 (NIPS
  2019)}, pages 5754--5764.

\end{thebibliography}
\bibliographystyle{acl_natbib}

\clearpage
\appendix

\begin{figure*}[t]
    \centering 
    \includegraphics[width=0.9\textwidth]{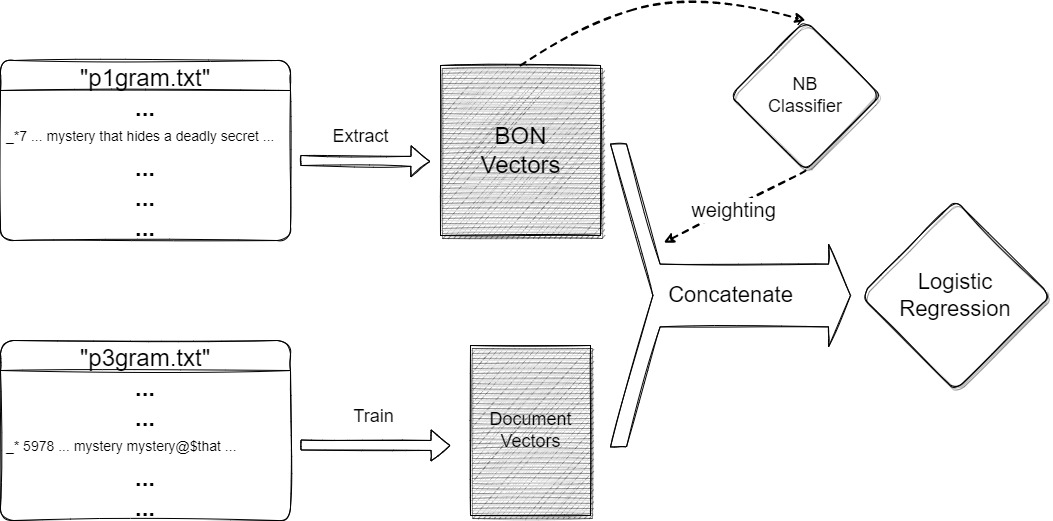} 
    \caption{Diagram of the ensemble and its input in the original code}
    \label{fig:ens_input}
\end{figure*}

\section{Detailed Description of the Bug}
\label{sec:appendix_bug}
As depicted in Fig.~\ref{fig:ens_input}, In the original work, neither of the two sets of vectors (DV-ngrams-cosine and NB-weighted BON) were obtained directly from the original IMDB movie review datasets. Instead, they were from two preprocessed versions of the IMDB movie review dataset stored in two files named "alldata-id\_p1gram.txt" and "alldata-id\_p3gram.txt", respectively. (We will refer to them as "p1gram.txt" and "p3gram.txt" in the remaining of this paper for short.) The file "p1gram.txt" contains unigrams extracted from the IMDB dataset, while "p3gram.txt" contains unigrams, bigrams and trigrams.

The problem is that the document orders in these two files are not the same. For example, the 7th document in "p1gram.txt" corresponds to the 5978th document in "p3gram.txt".

\begin{figure*}[t] 
    \centering 
    \includegraphics[width=0.9\textwidth]{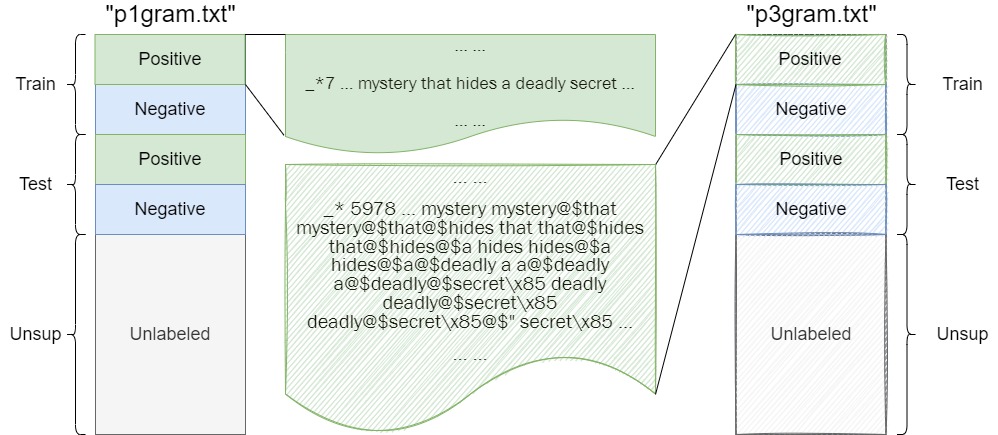} 
    \caption{Two preprocessed versions of the IMDB movie review dataset: p1gram.txt and p3gram.txt}
    \label{fig:ppd}
\end{figure*}

After a closer look at these two files, as shown in Fig.~\ref{fig:ppd}, one can see that in both files, the preprocessed documents are ordered in such a way that they can be naturally divided into blocks. Each block contains all the documents within one particular class and one particular subset (train, test or unlabelled). The orders of these blocks inside the two files are the same. For example, in both files, the positive examples from the training set occupy the first 12500 lines. 

However, inside each block, the documents are ordered differently in these two files. As a result, the document order of "p3gram.txt" is one in-block permutation (permutation within each block) away from the document order of "p3gram.txt".

In the original code, the DV-ngrams-cosine embeddings are built from "p3gram.txt", while the NB-weighted BON vectors are built from "p1gram.txt". They are directly concatenated without any reordering. This leads to the incorrect estimation of the test accuracy of the ensemble.

\section{Re-Evaluation of the Ensemble}
\label{sec:appendix_reev}
After finding the aforementioned bug, the next step is evaluating the ensemble correctly. Apart from re-evaluating the ensemble with the correct matching (so that the two vector representations of the same document are concatenated together), we also tried different ways of combining two documents by performing shuffling on one of the two files. Concatenating vectors from different documents is perfectly legit if no test labels are required in this process. Therefore, we compare different shuffling schemes, both using and not using test labels, for completeness and sanity check.

\subsection{Experiments}
\begin{figure*}[t]
    \centering 
    \includegraphics[width=0.9\textwidth]{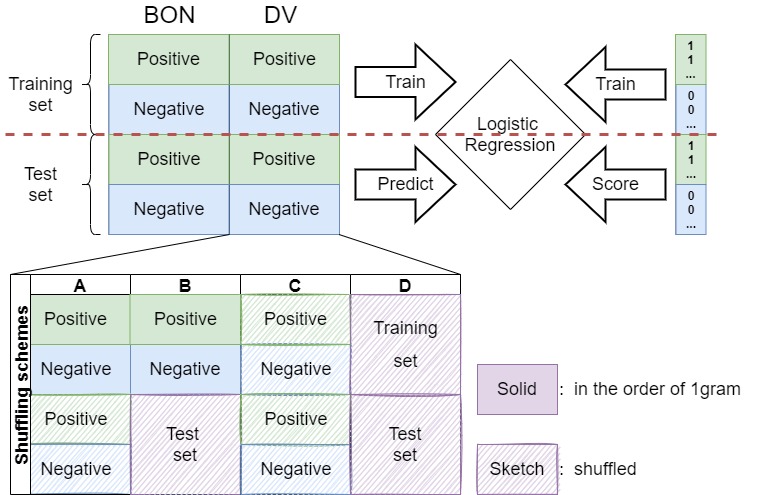} 
    \caption{The shuffling schemes for re-evaluation of the ensemble. First of all, both the Document Vectors (DV) and the BON vectors are sorted in the document order of p1gram.txt. Then, 4 shuffling schemes are imposed on the Document Vectors, respectively. In A and C, the blocks are shuffled internally, while in B and D, the corresponding positive block and negative block are mixed and shuffled as a whole. The train-test split is respected throughout this experiment. Neither of the BON vectors nor the labels are shuffled.}
    \label{fig:sf_ens}
\end{figure*}

The ensemble was evaluated by the original code, with both the original matching (the two representations were concatenated according to the document orders of the files "p1gram.txt" and "p3gram.txt") and the correct matching (the representations of the same document are concatenated). Some additional tests with different shuffling schemes were also carried out, only "p1gram.txt" was shuffled in all of these tests. The shuffling schemes are shown in Fig.~\ref{fig:sf_ens}. 

In both A and C, the test set was shuffled in-class. This requires using the labels of the test set in order to group documents with the same labels, which may significantly simplify the classification task. Both B and D only underwent cross-class shuffles, in which samples from different classes were mixed and treated equally. This augments the BON vectors with the DV-ngrams-cosine vectors of some random examples, thus, we basically add noise. In C this noise is only added to the test set, while in D it is also added to the training set, so the model can learn to ignore it. 

By comparing A vs. B and C vs. D, we can find out whether the aforementioned leakage of the test labels is necessary to obtain high test accuracy. In particular, C was designed to reproduce the incorrect evaluation in the original paper, except for the randomness in the shuffling process. All tests involving shuffling were run 30 times. 

\begin{table}
\centering
\begin{tabular}{lcc}
\hline
\textbf{Shuffling} & \textbf{Acc.} & \textbf{Acc.}\\
\textbf{Scheme} & \textbf{Mean} & \textbf{Std.}\\
\hline
\textbf{original matching} & 97.42 &  \\
\textbf{correct matching} & 93.68 &  \\
\textbf{test set in-class (A)} & 96.58 & 0.07 \\
\textbf{test set cross-class (B)} & 61.80 & 0.25 \\
\textbf{train/test in-class (C)} & 97.43 & 0.08 \\
\textbf{train/test cross-class (D)} & 91.64 & 0.08 \\
\hline
\end{tabular}
\caption{Test accuracy for different shuffling schemes.}
\label{tab:re}
\end{table}

\begin{figure}[t]
    \centering 
    \includegraphics[width=0.45\textwidth]{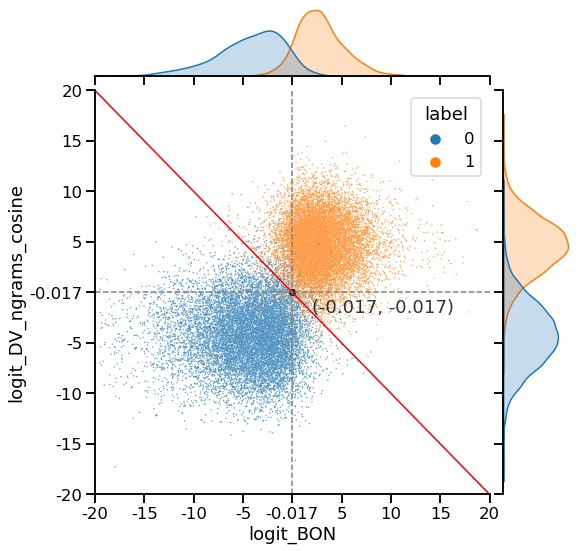}
    
    (a)
    
    \includegraphics[width=0.45\textwidth]{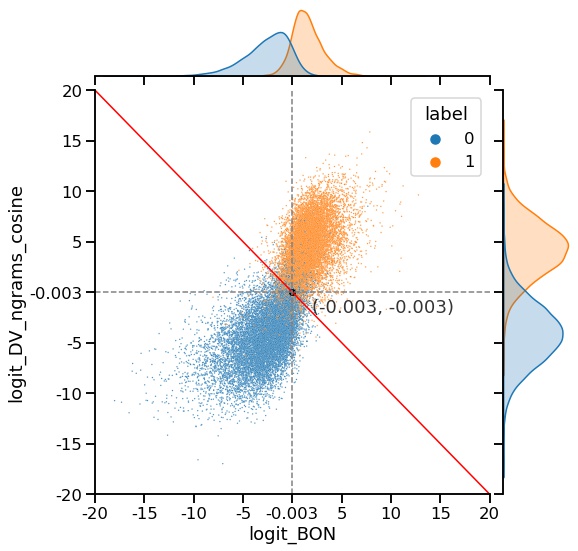}
    
    (b)
    
    \caption{Distributions of the logits (test set). The classification boundaries of the ensemble and the sub-models are noted as the red diagonal and the grey dashed lines, respectively. Here we evenly split the intercept of the ensemble for the 2 sub-models. (a) when the document vectors are incorrectly concatenated (as in the original paper); (b) when the document vectors are correctly concatenated.}
    \label{fig:logit_dist}
\end{figure}

\begin{figure}[h]
    \centering 
    \includegraphics[width=0.45\textwidth]{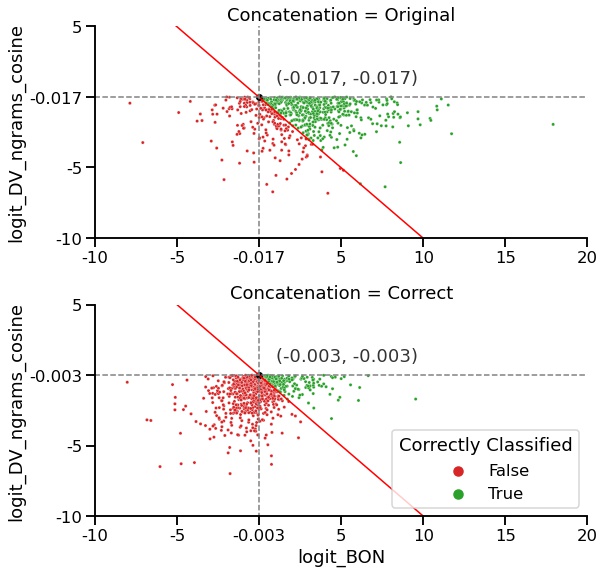}
    
    (a)
    
    \includegraphics[width=0.45\textwidth]{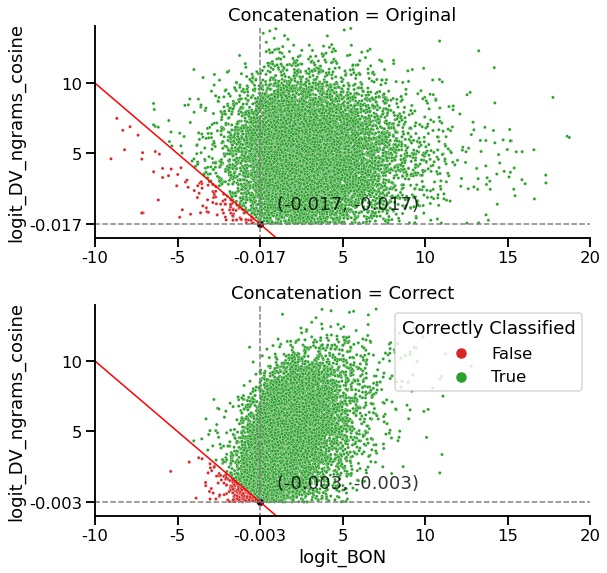}
    
    (b)
    
    \caption{Distributions of the logits of the positive examples. Green/red dots represent documents that are correctly/incorrectly classified by the ensemble. (a) positive documents that are misclassified by the DV\_ngrams\_cosine part, (b) positive documents that are correctly classified by the DV\_ngrams\_cosine part}
    \label{fig:logit_dist_zoom}
\end{figure}

\begin{figure}[h]
    \centering 
    \includegraphics[width=0.45\textwidth]{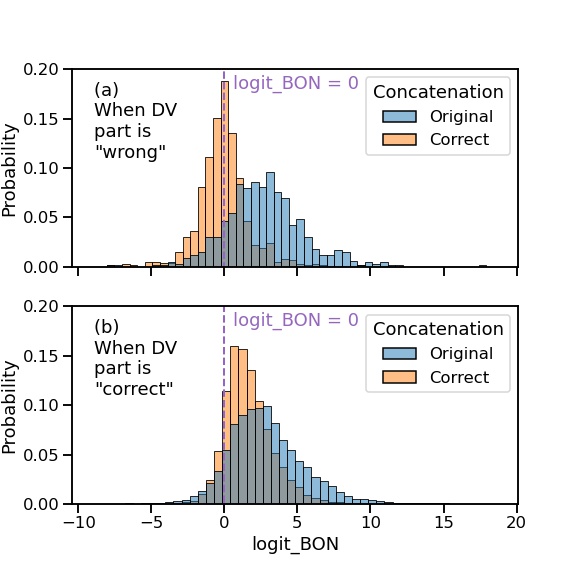}
    \caption{Marginal distributions of the BON part of the positive examples. (a) positive documents that are misclassified by the DV\_ngrams\_cosine part, (b) positive documents that are correctly classified by the DV\_ngrams\_cosine part}
    \label{fig:logit_dist_marg}
\end{figure}

\subsection{Results and discussions}
The results of the shuffling tests are shown in Table~\ref{tab:re}. With the correct matching, the test accuracy of the ensemble was 93.68\%, which is much lower than the test accuracy of 97.42\% achieved by the original matching (this result agrees with the original paper).

Augmenting the BON vectors with the DV-ngrams-cosine embeddings of different documents from the same classes (A and C) showed much higher performance than the correct matching. Between them, C got better accuracy due to the consistency in the training and testing process. On the other hand, the training examples for A were correctly concatenated so the model suffered from data drift during testing.

Augmentation using some random documents for the test set only (B) makes the logistic regression perform almost like a random classifier, while when it is done for the training set also (D), it can learn to ignore the added noise. They serve as a sanity check to verify that shuffling without knowing the labels doesn't get better accuracy than the "correct matching".

Only the shuffling scheme C yielded accuracy that is very similar to the original matching, showing that a random in-class shuffle of both the training and test set can closely reproduce the test accuracy reported in the original paper.

\subsection{A closer look at the high estimation of accuracy caused by the incorrect concatenation}
\label{sec:subsec_hard_n_easy}
From the accuracy of each part of the ensemble (table \ref{tab:imdb_acc}), we see that each part of the ensemble is incorrect for a rather small portion of examples (less than 7\% for DV-ngrams-cosine, less than 9\% for NB-weighted BON). As can be seen from the marginal distributions in figure \ref{fig:logit_dist}, for these incorrectly predicted examples the absolute values of logits are usually much smaller compared to the correctly classified examples. 

We can also see from the scatter plot of figure \ref{fig:logit_dist} (a) that when the vectors are concatenated in the original order, the logits of the two parts (DV-ngrams-cosine and NB-weighted BON) look independent conditioned on the true label, which corroborates our speculation that "p3gram.txt" were randomly shuffled inside each class. 

Thus, there is a high chance that for an incorrect prediction by the DV-ngrams-cosine part ("hard" example) a random example of the same class will have larger NB-weighted BON logit with the correct sign ("easy" example) and fix this prediction. But as shown in the scatter plot of figure \ref{fig:logit_dist} (b), when the vectors are correctly concatenated, the logits of the two parts have a positive correlation. So, the correct model doesn't benefit from the aforementioned independence.

To see it more clearly, Figure \ref{fig:logit_dist_zoom} (a) shows only the examples that are incorrectly classified by the DV-ngrams-cosine part. In the original combination, the complementary BON logit is mostly positive and quit large, moving many of those examples to the correct side of the decision boundary. When concatenated correctly, the BON logit is concentrated near zero and rarely helps. In Figure \ref{fig:logit_dist_zoom} (b) only those examples that are correctly classified by the DV-ngrams-cosine part are shown. We can see that adding the BON logit can sometimes move those examples to the incorrect side of the boundary. However, the proportion of such cases is similar for the original and correctly concatenated representations.

Figure \ref{fig:logit_dist_marg} shows the marginal distributions of the NB-weighted BON part in the same settings. Between the two situations (the DV-ngrams-cosine part is incorrect/correct), the histogram of the BON logits stays in a very similar shape when the vectors are incorrectly concatenated (blue), while showing a shift when the vectors are correctly concatenated (orange). These observations corroborate the argument that the model on the incorrectly concatenated vectors benefits from the aforementioned conditional independence.

It is also worth noticing from figure \ref{fig:logit_dist_marg} that in both (a) and (b), the distribution of the BON logits of the original concatenation (blue) is more on the positive side. This indicates that incorrect concatenation may also help the training of the weights for NB-weighted-BON. We have tentatively run tests on this, but haven't found any overall improvements: the ensemble trained on incorrectly concatenated vectors and tested on correctly concatenated vectors only has a test accuracy of ~93.0\%.

\section{Training DV-ngrams-cosine with NB Sub-Sampling}
\label{sec_appendix_subsamp}
In the experiments of this section (also in section \ref{sec:subsamp}), the train/validation/test splits \citep{suchin2020} of the IMDB dataset were also used. During the training, after every epoch, logistic regression was trained on the training set; the hyperparameter $C$ was tuned on the validation set; then the model with the "best" $C$ was tested on the test set. The training of each type (with or without NB sub-sampling) was repeated 3 times.\newline

\begin{figure}[h] 
    \centering 
    \includegraphics[width=0.47\textwidth]{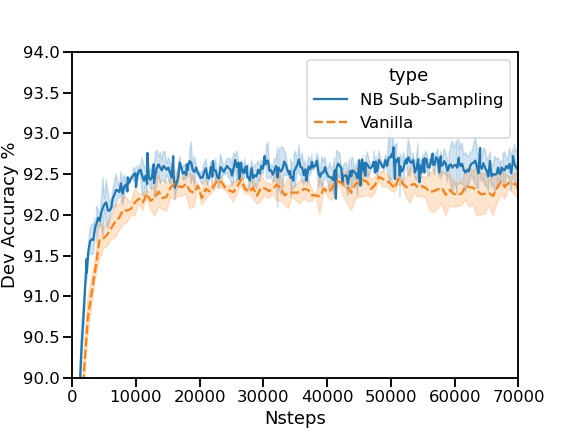}
    \caption{Training process with and without NB sub-sampling. The \textbf{validation} accuracy of the logistic regression built on top of the document vectors is plotted. The mean values and standard deviations were calculated over 3 runs for each type.}
    \label{fig:sub_samp_dev}
\end{figure}

The validation accuracy is plotted in Figure \ref{fig:sub_samp_dev} and the test accuracy is plotted in Figure \ref{fig:sub_samp}. The runs with NB sub-sampling showed slight advantages in both plots, almost everywhere.

\end{document}